\pdfoutput=1

\documentclass[11pt]{article}

\usepackage{paclic35}

\usepackage{svg}

\usepackage{multirow}
\usepackage{times}
\usepackage{latexsym}
\usepackage{graphicx}
\usepackage{algorithmic}
\usepackage{algorithm}
\usepackage{subcaption}
\usepackage{mathrsfs}
\usepackage{soul}
\usepackage{footmisc}
\usepackage{ragged2e}
\usepackage{booktabs}
\usepackage[normalem]{ulem}
\usepackage{balance}

\usepackage[T1]{fontenc}

\usepackage[utf8]{inputenc}

\usepackage{CJKutf8}
\usepackage{microtype}

\title{ParaCotta: Synthetic Multilingual Paraphrase Corpora 
\\ from the Most Diverse Translation Sample Pair}

\author{
    {\bf Alham Fikri Aji}$^{\ast\bullet}$\quad
    {\bf Tirana Noor Fatyanosa}$^{\ast\circ}$\quad
    {\bf Radityo Eko Prasojo}$^{\ast\star}$\quad
    {\bf Philip Arthur}$^{\dagger}$\quad \\
    {\bf Suci Fitriany}$^{\ast}$\quad
    {\bf Salma Qonitah}$^{\ast}$\quad
    {\bf Nadhifa Zulfa}$^{\ast}$\quad
    {\bf Tomi Santoso}$^{\ast}$\quad
    {\bf Mahendra Data}$^{\ddagger\circ}$\quad \\
    $^{\ast}$ Kata.ai Research Team, 
    $^{\circ}$ Kumamoto University,
    $^{\bullet}$ University of Edinburgh, \\
    $^{\star}$ Universitas Indonesia,
    $^{\dagger}$ Oracle Digital Assistant,
    $^{\ddagger}$ Brawijaya University\\
    \texttt{\{aji,tirana.fatyanosa,ridho,suci\}@kata.ai} \\ \texttt{\{salma.qonitah,dhifa.zulfa,tomi.santoso\}@kata.ai} \\
    \texttt{\{fatyanosa, mahendra.data\}@dbms.cs.kumamoto-u.ac.jp}\\
    \texttt{philip.arthur@oracle.com, mahendra.data@ub.ac.id}
}

\begin{document}
\maketitle
\begin{abstract}


We release our synthetic parallel paraphrase corpus across 17 languages: Arabic, Catalan, Czech, German, English, Spanish, Estonian, French, Hindi, Indonesian, Italian, Dutch, Romanian, Russian, Swedish, Vietnamese, and Chinese. 
Our method relies only on monolingual data and a neural machine translation system to generate paraphrases, hence simple to apply.
We generate multiple translation samples using beam search and choose the most lexically diverse pair according to their sentence BLEU.
We compare our generated corpus with the \texttt{ParaBank2}.
According to our evaluation, our synthetic paraphrase pairs are semantically similar and lexically diverse.




\end{abstract}

\section{Introduction}
Paraphrases are semantically similar sentences or phrases using different expressions~\cite{Bhagat2013}. A paraphrase generation system can be developed by training a model, given a dataset of paraphrase parallel texts~\cite{egonmwan2019}. Paraphrase parallel corpus is accessible in English~\cite{Dolan2005,fader2013,xu2015}. However, such data for other languages are not as common. \cite{ganitkevitch2014multilingual} proposed a multilingual paraphrase pairs dataset; however, their corpus is only on phrase-level.


By utilizing a machine translation system, \newcite{wieting2017paranmt} proposed synthetic paraphrase corpus by back-translating bilingual text. 
However, this approach does not consider lexical diversity for the generated paraphrases.
\newcite{EdwardHu2019} and \newcite{Hu2019} further improve the method by applying a constraint to generate more diverse paraphrases, then choosing diverse pairs as the synthetic dataset. These methods require bilingual corpus, which might not be easily accessible for certain languages or domains. 
Our work takes inspiration from selecting the most diverse pair; however, we remove the bilingual text requirement. 

We propose a simple way to generate paraphrases by selecting the most diverse pair (in terms of BLEU) from the translation sample. Our approach generates paraphrases from a monolingual text; therefore, not bound to the availability of parallel corpus. We also show that this technique can produce diverse paraphrases, measured in the BLEU score. In addition, we release our generated paraphrase dataset in 17 languages.

\begin{figure*}[ht]
\centering
 \includegraphics[trim=0 1300 180 0,clip,width=1.6\textwidth]{"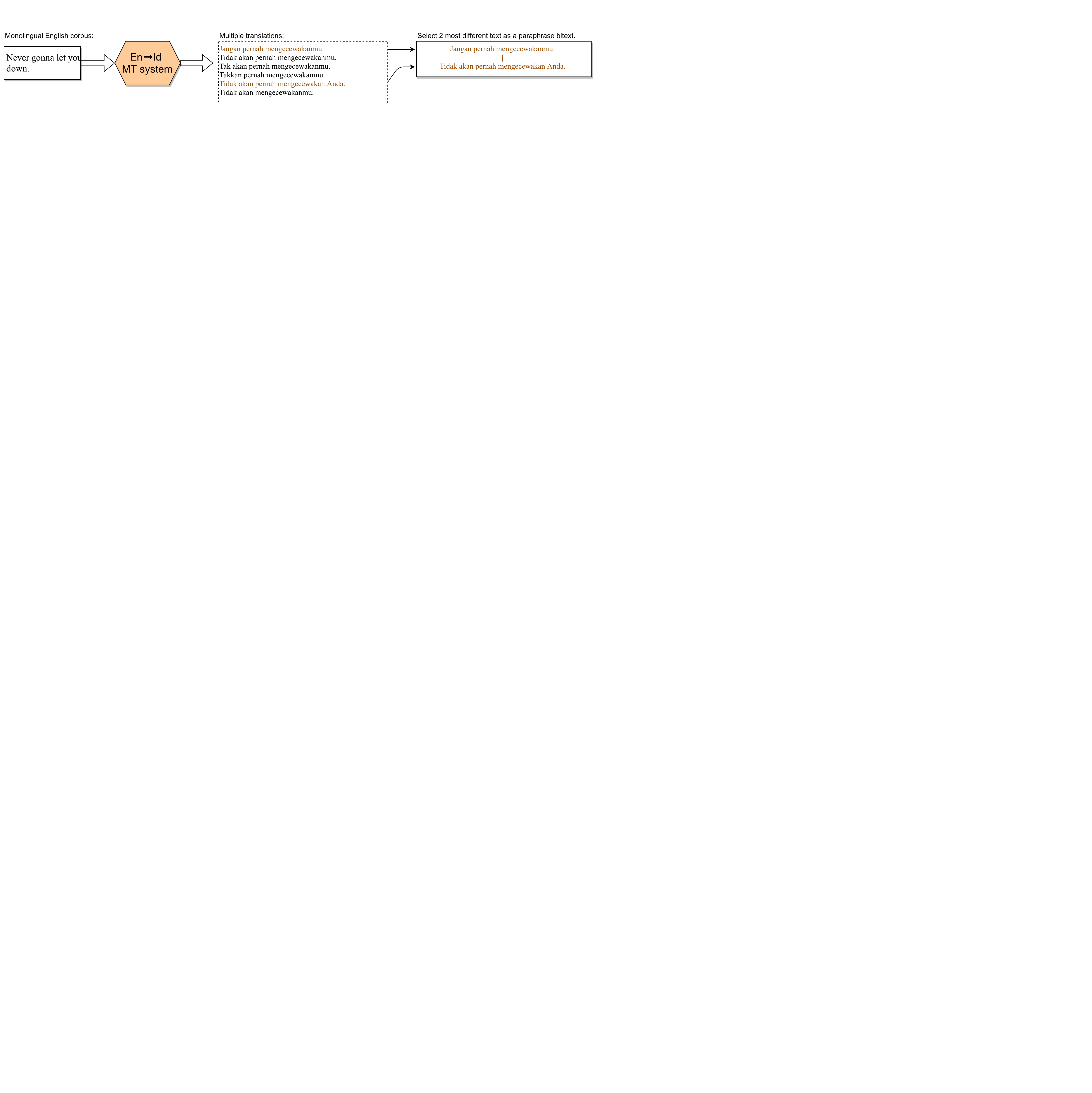"}
 \caption{An example of synthetic paraphrase corpus generation using a machine translation system.}
  \label{fig:bleu_selection}
\end{figure*}

\begin{table*}[ht!]
    \centering
    \small
    \resizebox{\textwidth}{!}{%
    \begin{tabular}{@{}l@{}c@{ ~ }c@{ }c@{}}
    \toprule
        Text & Sem.Similarity & \multicolumn{2}{c}{Lexical Diversity} \\
             & stsb cosine↑ & BLEU↓ & Jaccard↓     \\
    \midrule
        He's denied them protection. & 0.630 & 1.7 & 0.0 \\
        They're not allowed to do that in a protective shield. &  \\
        \midrule
        voter representation cannot be guaranteed. &  0.917 & 2.0 & 0.0 \\
        It is not possible to guarantee the right to vote. \\
        \midrule
        It is therefore necessary to compensate for business tax failures in the coming years.  & 0.796 & 6.9 & 0.273 \\
        Therefore, the trade tax losses would have to be compensated in the next few years. &  \\
        \midrule
        Therefore, unavoidable waiting times may occur.  & 0.866 & 10.7 & 0.250 \\
        For this reason, there may be inevitable waiting times. &  \\
        \midrule
        Maintenance-free batteries are supposed to prevent this from happening. & 0.921  & 16.9 &  0.308 \\
        Maintenance-free batteries should actually prevent that.  \\
        \midrule
        Do taxes need to be raised to finance the stimulus package?& 0.949  & 21.0 &  0.615\\
        Do taxes have to be increased to finance the economic stimulus package? \\
        \midrule
        Small successes for the first time with tested Corona vaccine (9.45 o'clock) & 0.959  & 38.6 & 0.533\\
        Small successes with tested Corona vaccine (9.45 am) \\
        \midrule
        Everything is now clear for the construction of a new ice channel at Barenberg. & 0.994 & 43.6 & 0.812 \\
        Now everything is clear for the start of construction of a new ice channel on Barenberg. \\

        \bottomrule
    \end{tabular}}
    \caption{Synthetic paraphrase corpus example (English)}
    \label{tab:english-result}
\vspace{-18px}
\end{table*}




\section{\textls[-18]{Generating Paraphrase via Diverse Pairs}}




We propose a way to construct synthetic paraphrase corpus by utilizing a machine translation system. Our approach involves translating texts from English to the desired language, therefore not limited to the availability of bilingual corpora for back-translation~\cite{EdwardHu2019}. Specifically, given an input text $X$, we produce several translation samples $Y_0, ..., Y_N$ with beam-search. Then, we chose two sentences $Y_i$ and $Y_j$ as a paraphrase pair, such that both sentences are the most lexically diverse among other choices. Here, we define the lexical diversity with a BLEU score, where a lower BLEU score denotes a more diverse pair. For more details, see Figure \ref{fig:bleu_selection}.
\vspace{-2px}

To produce the synthetic paraphrase corpus for a language $L$, we use an English to $L$ translation system, as well as monolingual English corpus. It is possible to use a pivot language other than English. However, we argue that it is more difficult to achieve due to the availability of the translation system.
\vspace{-2px}

With this method, we generate synthetic paraphrase corpus across 17 languages. For non-English corpus, we translate monolingual English to the desired language. Our English monolingual corpus is sampled from ParaBank2 \cite{Hu2019} (3M sent), Wikipedia (1M sent), NewsCrawl (1M sent), and  English Tatoeba (1M sent). For the English paraphrase corpus, we translate monolingual German text collected from NewsCrawl (2.5M sent) and German Tatoeba (500k). We are planning to support more languages and use more monolingual data as future work.
Examples of English-generated data can be seen in Table~\ref{tab:english-result}, alongside their qualitative evaluations, which will be explained in Section~\ref{sec:eval}.
\vspace{-1.6px}


\section{Model Configuration}
\vspace{-1.7px}
We use a Transformer-based encoder-decoder architecture~\cite{Vaswani2017}  for both our translation system and paraphrase generator system. For both systems, we use the same Transformer-base architecture which consists of 6 layers of encoder and decoder, and an embedding size of 512. The input is tokenized with sentence-piece~\cite{kudo2018sentencepiece}. 

We rely on NMT system to produce our synthetic paraphrase data.
For most of our translation system, we use pre-existing public model available in Huggingface.\footnote{https://huggingface.co/Helsinki-NLP} These MT systems are trained on OPUS parallel corpus. Without losing the generality, we re-train Indonesian MT with the additional dataset from~\newcite{guntara2020benchmarking}.

Similarly, we use the same architecture to train our paraphrase generation model. We train our paraphrase system for 10 epochs with Adam optimizer. We use Marian toolkit~\cite{mariannmt} to train our model.

\section{Evaluation and Analysis}
\label{sec:eval}

\subsection{Evaluation Method}

Our objective is to maximize both the lexical diversity and semantic similarity of our paraphrase pairs. The lexical diversity is, by the design of our approach, guarded by the BLEU score. Indeed, using the BLEU score to determine paraphrase originality or diversity has been used in prior work~\cite{Mallinson2017,Hu2019,EdwardHu2019}. However, in our case, we use BLEU exclusively to measure lexical diversity. On top of BLEU, we further evaluate our paraphrase quality using word-level Jaccard Index~\cite{jaccard1912distribution} for lexical diversity. For semantic similarity, we rely on using manual evaluation and and sBERT score~\cite{reimers2020making}.

We average the BLEU score for both directions since the reference paraphrase is not defined.
Following~\cite{Hu2019}, we also compute the BLEU on lowercased text after stripping the punctuation. We use sacreBLEU~\cite{post2018call} for calculation.
Similarly, we compute the Jaccard index on lowercased and de-punctuated text.

For automatic semantic similarity evaluation, we leverage distilled multilingual sBERT models~\cite{reimers2020making}, in particular the \texttt{paraphrase-xlm-r-multilingual-v1} and \texttt{stsb-xlm-r-multilingual} models trained on 50+ languages, which scored a high Pearson's $\rho$ on semantic textual similarity (STS) tasks despite being relatively lightweight.

For manual evaluation, we randomly select 100 sentences from the generated corpus. Then, professional annotators\footnote{This is to control the annotation quality better and to avoid navigating through the ethical concerns of using a crowd-platform~\cite{shmueli2021beyond}. However, this limits our manual evaluation to only two languages in which our annotators are professionally fluent.} are asked to score each of the paraphrase pairs on a 3-point Likert scale system: (1) Inequivalent or unrelated; (2) Roughly equivalent; (3) Completely or mostly equivalent. The scores are then averaged and scaled to 0-100. A more detailed guideline can be found in Appendix~\ref{sec:manualevalguide}.

\begin{table}[t!]
\small
    \centering
    \resizebox{\columnwidth}{!}{%
    \begin{tabular}{@{ }lc@{ }c@{ }c@{ }c@{ }}
    \toprule
                & \multicolumn{2}{c}{Semantic Similarity} & \multicolumn{2}{c}{Lexical Diversity} \\
        Language & stsb↑ & para↑ &  BLEU↓ & Jaccard↓ \\
        \midrule
Arabic	(ar)	&	0.926	& 0.925 & 25.6	&	0.357	\\
Catalan	(ca)	&	0.909	& 0.901 & 34.3	&	0.435	\\
Czech	(cs)	&	0.913	& 0.923       & 24.7	&	0.376	\\
German	(de)	&	0.934	&   0.925   & 28.0	&	0.427	\\
English	(en)	&	0.909	& 0.876	&   34.6	&	0.523	\\
Spanish	(es)	&	0.942	&	0.932   & 34.0	&	0.452	\\
Estonian	(et)	&	0.892	&	0.911 & 23.2	&	0.377	\\
French	(fr)	&	0.924	& 0.914 & 	33.3	&	0.425	\\  
Hindi	(hi)	&	0.894	& 0.897 &	39.5	&	0.604	\\
Indonesian	(id)	&	0.936	& 0.929 & 	28.1	&	0.426	\\
Italian	(it)	&	0.931	& 0.920 &	31.6	&	0.421	\\
Dutch	(nl)	&	0.921	& 0.912 &	30.4	&	0.456	\\
Romanian	(ro)	&	0.933	& 0.927 & 	26.9	&	0.376	\\
Russian	(ru)	&	0.930	& 0.921 & 	26.9	&	0.376	\\
Swedish	(sv)	&	0.916	& 0.906	& 29.2	&	0.428	\\
Vietnamese	(vi)	&	0.933	& 0.904	& 40.2	&	0.517	\\
Chinese	(zh)	&	0.879	& 0.877		& 37.8 & 0.470		\\

\bottomrule
    \end{tabular}}
    \caption{Corpus statistic across languages. stsb and para are the cosine distance of the embeddings generated by sBERT \texttt{stsb} and \texttt{paraphrase} models, respectively.}
    \label{tab:all-lang}
\vspace{-15px}
\end{table}

\subsection{Synthetic Corpus Evaluation}

\begin{table*}[ht!]
    \centering
    \small
    \resizebox{\textwidth}{!}{%
    \begin{tabular}{lcccc|cccc}
    \toprule
         Dataset & \multicolumn{2}{c}{Semantic Similarity} &  \multicolumn{2}{c}{Lexical Diversity} & \multicolumn{2}{c}{Semantic Similarity} &  \multicolumn{2}{c}{Lexical Diversity}  \\
         & Manual↑ & Cosine↑ & BLEU↓ & \multicolumn{1}{c}{Jaccard↓} & \multicolumn{1}{c}{Manual↑} & Cosine↑ & BLEU↓ & Jaccard↓ \\
         \midrule
         & \multicolumn{4}{c}{\textbf{English dataset}} &  \multicolumn{4}{|c}{\textbf{Indonesian dataset}} \\
         ParaBank2 \cite{Hu2019} & 88.5 & 0.812 & \textbf{23.9} & \textbf{0.388} & \multicolumn{4}{c}{n/a} \\
         Ours (no filter) & 95.0 & 0.876 & 34.6 & 0.523 & 92.5	&	0.936 &	28.1 & 0.426 \\
         Ours (BLEU filter 0-80) & 95.0 & 0.909	& 34.1 & 0.522 & 92.3 & 0.936 &	28.1 & 0.426 \\
         Ours (BLEU filter 0-60) & 94.8 & 0.908	& 31.7 & 0.512 &  91.2	& 	0.935	& \textbf{26.4} & \textbf{0.420} \\
         Ours (BLEU filter 20-80) & \textbf{97.2} & \textbf{0.926} &41.7 & 0.594 & \textbf{96.6}	&	\textbf{0.953}	& 40.5 & 0.566 \\
         Ours (BLEU filter 20-60) & 97.0 & 0.924	& 39.2 & 0.585 & 95.2	&	0.952 &	38.5 & 0.573 \\
         \bottomrule
         
    \end{tabular}}
    \caption{Corpus statistic with human evaluation.}
    \label{tab:data-eval}
\vspace{10px}
\end{table*}


The data statistic across 17 languages, sampled from 10k sentences per language, can be seen in Table~\ref{tab:all-lang}. We find that the scores on both models are extremely similar; therefore, we only used the \texttt{stsb} model for our later evaluations.

Table~\ref{tab:mutilang_example} shows some examples on our proposed dataset in other languages besides English.
\begin{table*}[ht!]
    \centering
\includegraphics[trim=0 45 0 0,width=0.98\textwidth]{"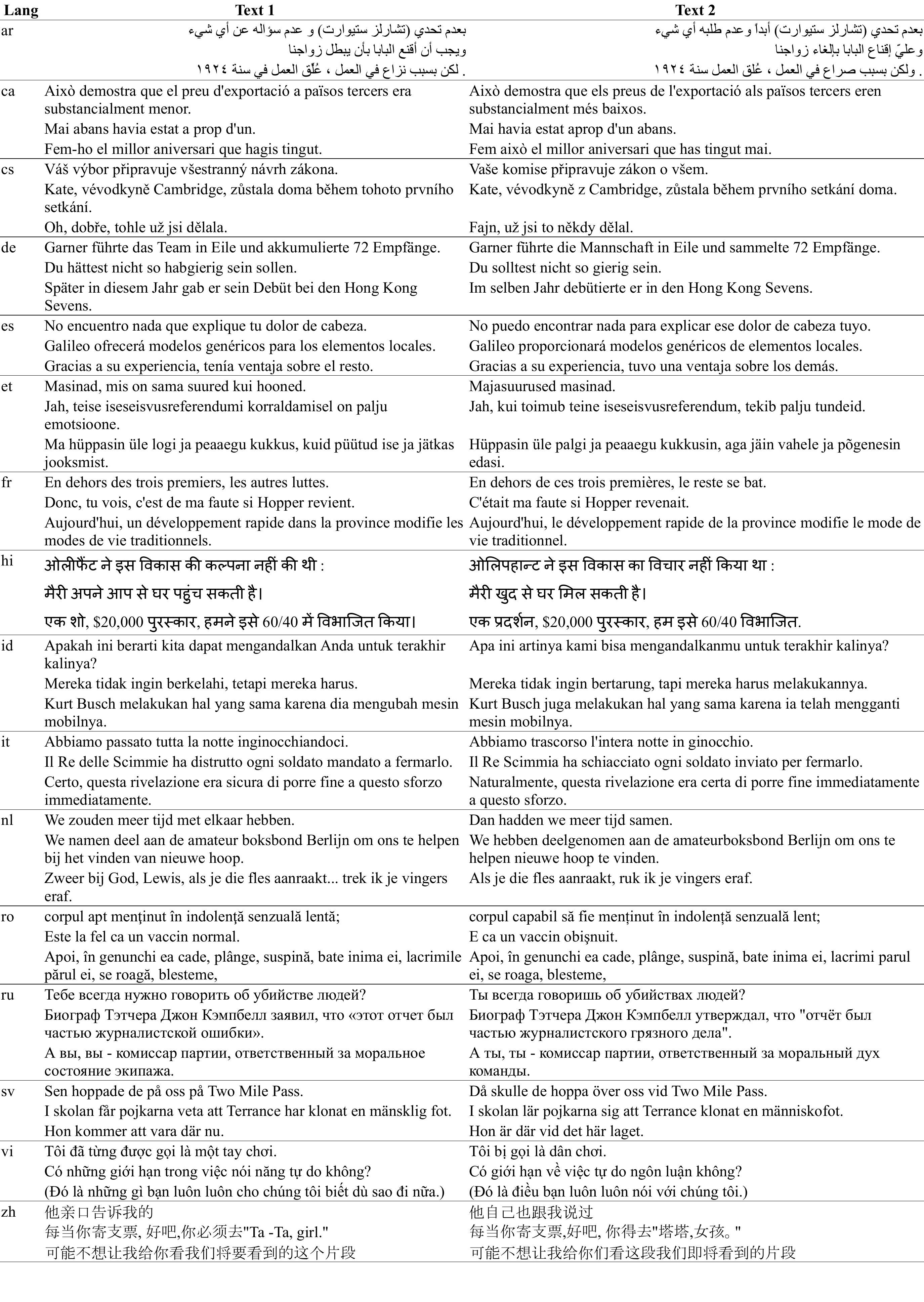"}
    \caption{ParaCotta example on other languages.}
    \label{tab:mutilang_example}
\end{table*}

To further analyze our generated dataset, we perform human evaluation on selected languages of English and Indonesian. We also evaluate English ParaBank2 as a comparison.
We manually annotate 100 samples from our dataset per language. For ParaBank2, we manually annotate 50 samples. We achieve an annotator agreement of 0.5 (weighted kappa) which indicates fair agreement.

As shown in Table~\ref{tab:data-eval}, our proposed technique is able to generate semantically similar paraphrases. Compared to ParaBank2, we achieve a better semantic similarity score. Unfortunately, our dataset is less lexically diverse. However, our approach uses a monolingual corpus to produce the paraphrase data. Therefore, our approach does not depend on the availability of parallel corpus, which is beneficial for low-resource languages. Note that our approach still requires parallel corpus to build the MT system, although an alternatively zero-shot MT system can be used. Similarly, MT system can be build under low-resource setting with the help of pre-trained language models. In these cases, our paraphrase generation mechanism can be used regardless the availability of the parallel corpus. However, we leave this as future work.


\label{sec:pairselect}

\subsubsection*{Metric Correlation}

To test the relationship between all metrics for all models, we calculated the Spearman correlation with $\alpha = 0.05$ as shown in Table~\ref{tab:spearman}.
BLEU and Jaccard correlate well to measure lexical diversity, with a 0.803 Spearman coefficient. Similarly, human evaluated semantic similarity correlates with sBERT cosine similarity with 0.304 Spearman coefficient.

\begin{table}[]
\centering
    \small
\caption{Spearman correlation between all scores for all models (Indonesian dataset).}
\label{tab:spearman}
\resizebox{0.48\textwidth}{!}{
\begin{tabular}{l|cccc}
\cline{2-5}
\multicolumn{1}{c|}{} & Manual & Cosine & BLEU & Jaccard \\ \hline
Manual & 1 & 0.304 & 0.210 & 0.233 \\
Cosine & 0.304 & 1 & 0.465 & 0.605 \\
BLEU & 0.210 & 0.465 & 1 & 0.803 \\
Jaccard & 0.233 & 0.605 & 0.803 & 1 \\ \hline
\end{tabular}}
\end{table}

\begin{figure}[t]
	\centering
	\subfloat[Manual vs Cosine]{
	    \centering
		\fbox{\includegraphics[height = 2.5cm,width = 7.5cm]{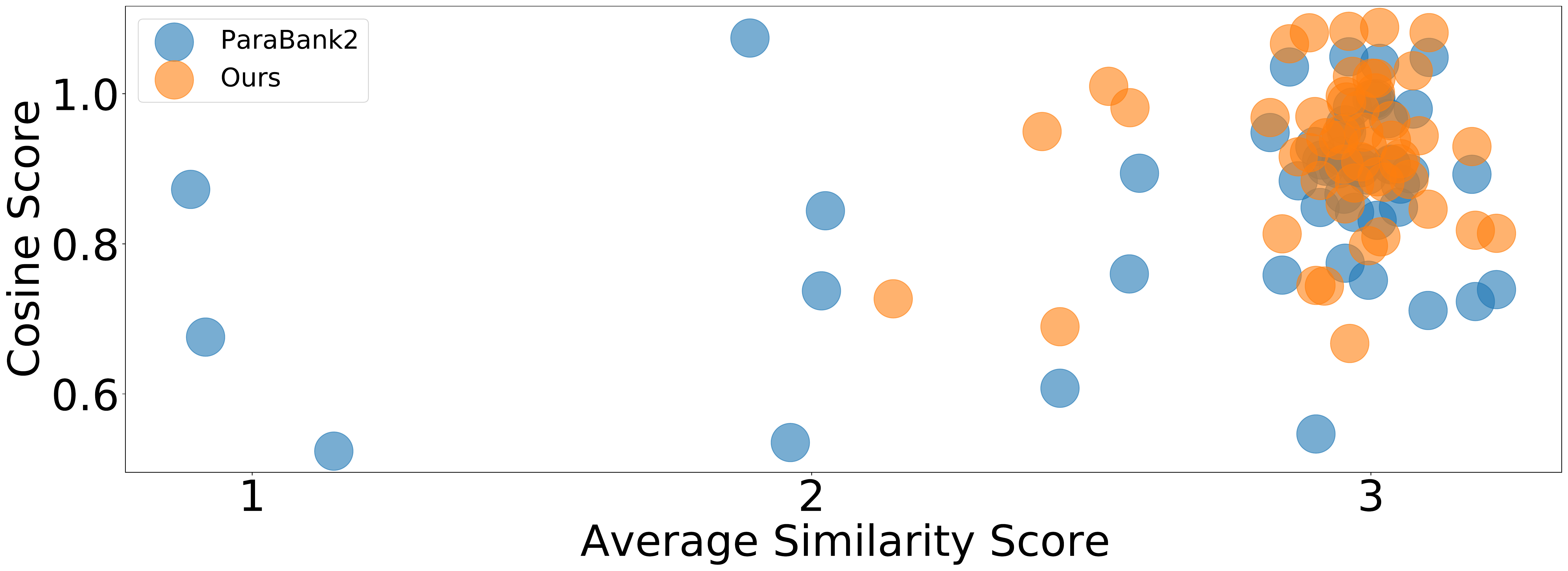}}
	}
	\quad
	\subfloat[BLEU vs Jaccard]{
	    \centering
		\fbox{\includegraphics[height = 2.5cm,width = 7.5cm]{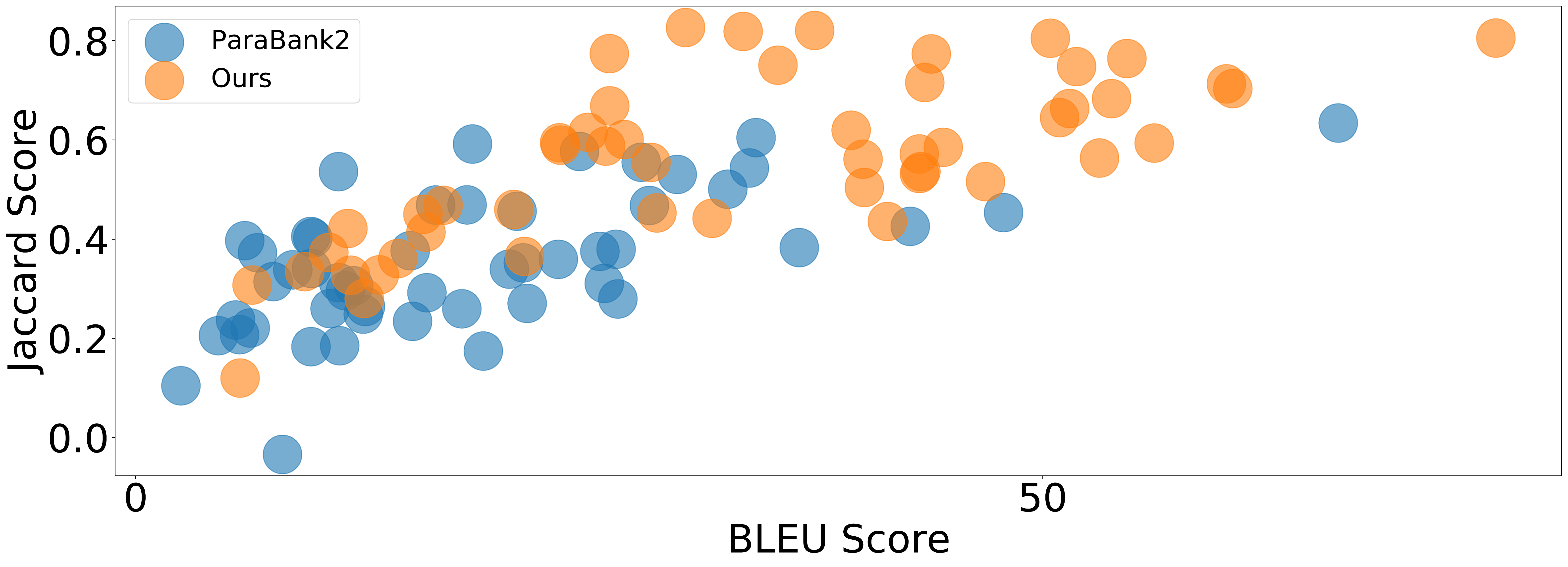}}
	}
	\caption{Scatter plots comparison between ParaBank2 and Ours for metrics correlation. Random gaussian noise $\mathcal{N}$(0,0.1) and $\mathcal{N}$(0,0.05) have been added to x-axis and y-axis, respectively. 
	} \label{fig:scatterplot}
\end{figure}

In the scatter plots (Figure~\ref{fig:scatterplot}), we visualize the comparison between ParaBank2 and our approach for the English dataset. We can see that all the metrics have a higher density on high-scoring sentences as most of the sentences are given a score of 3 by human annotators. Overall, our proposed approach is not as diverse as ParaBank2 but generally has a higher human annotation value than ParaBank2.

\subsubsection*{BLEU-filtering}


\begin{figure}[t]
	\centering
	\subfloat[English dataset]{
	    \centering
		\fbox{\includegraphics[height = 2.5cm,width = 7.5cm]{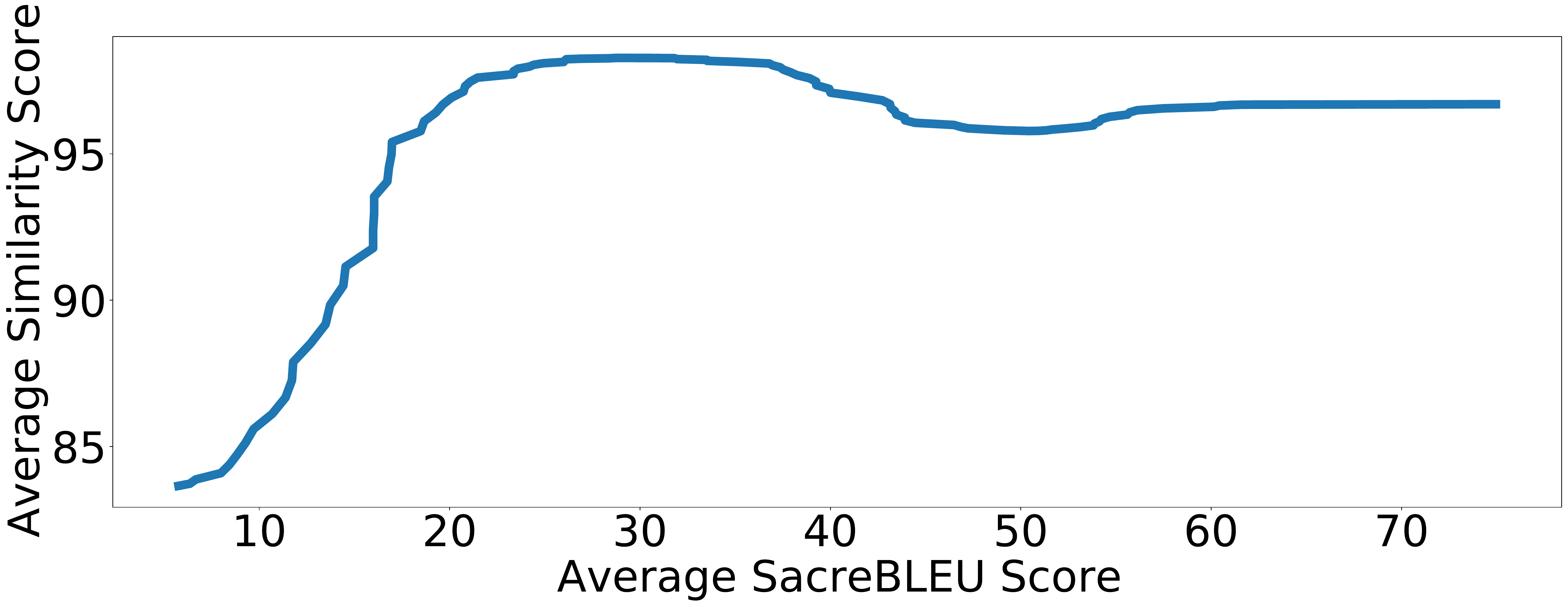}}
	}
	\quad
	\subfloat[Indonesian dataset]{
	    \centering
		\fbox{\includegraphics[height = 2.5cm,width = 7.5cm]{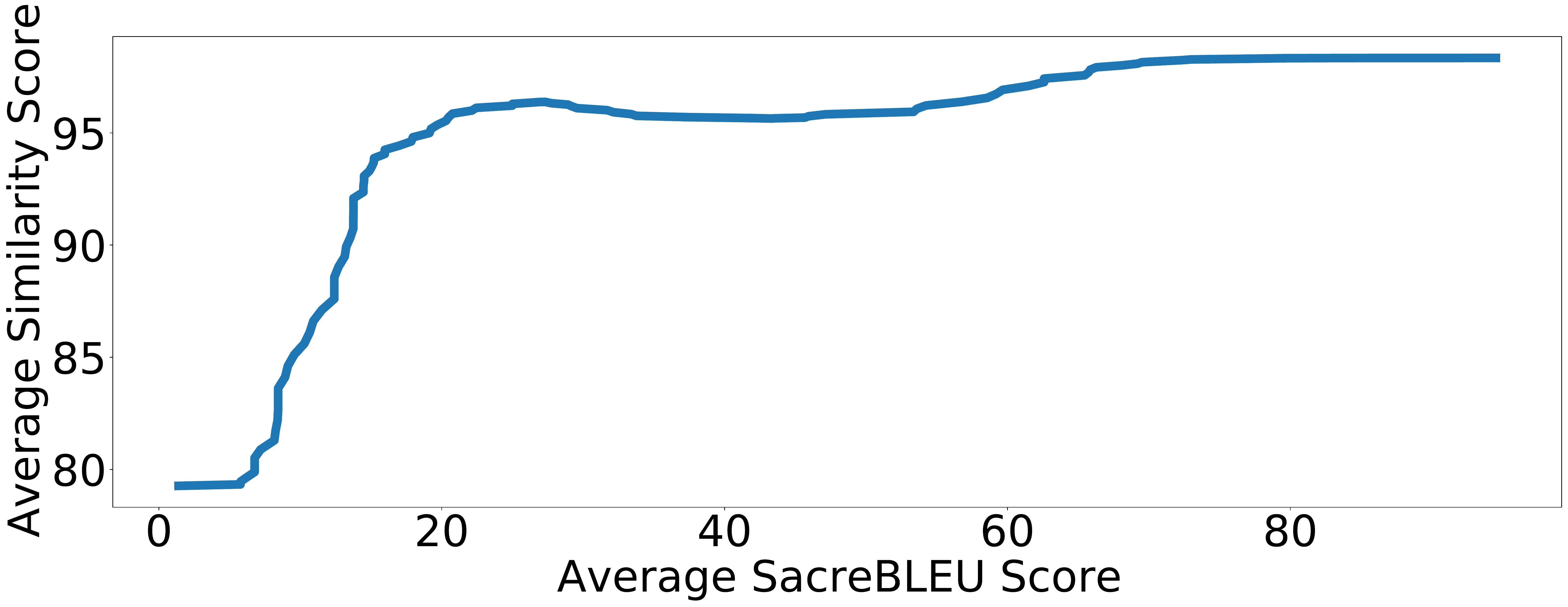}}
	}
	\caption{Comparison of BLEU score and manual semantic similarity score.}\label{fig:correlation}
\end{figure}

Upon further investigation, we notice a correlation between the BLEU and human-annotated semantic similarity. As shown in Figure~\ref{fig:correlation}, less creative paraphrases tend to be more semantically similar. In contrast, more diverse paraphrases are less semantically similar. Based on this observation, we also attempt to filter our generated synthetic pairs based on BLEU. Specifically, high-BLEU paraphrases can be removed to avoid `lazy and boring' paraphrases, resulting in more diverse datasets. Orthogonally, low-BLEU paraphrases can be removed as they are not as semantically similar. Filtering our corpus can further adjust the overall lexical diversity and semantic similarity, as shown in Table~\ref{tab:data-eval}.

\subsection{Paraphrase System Evaluation}
\begin{table}[ht!]
\centering
\label{tab:results}
\small
\begin{tabular}{@{}l@{ }cc@{}c@{}}
\toprule
\multicolumn{1}{c}{\multirow{2}{*}{Model}} & \multicolumn{2}{c}{Semantic Similarity} & Diversity \\ 
\multicolumn{1}{c}{} & Manual↑ & Cosine↑ & BLEU↓ \\ 
\midrule
Round-trip MT & 78.1 & 86.8 & 44.7 \\
Ours (no filter) & 83.5 & 86.2 & \textbf{32.8} \\
Ours (BLEU filter 20-60) & 88.1 & 91.2 & 47.0 \\
Ours (BLEU filter 20-80) & \textbf{88.9} & \textbf{92.1} & 48.0 \\ \bottomrule
\end{tabular}
\caption{Automatic and manual evaluation on 100 sentences across models for single reference (Indonesian dataset).}
\label{tab:results}
\end{table}

In this section, we build our paraphrase system by training a Transformer seq2seq model~\cite{Vaswani2017} with our proposed synthetic paraphrase corpus. As another comparison, we also implement round-trip MT, where we translate the input to a pivot language and then translate it back to input language~\cite{Mallinson2017}.


\vspace{10px}
Table \ref{tab:results} shows the overall results for lexical diversity and semantic similarity.
From the result, Round-trip MT achieves the least semantically similar paraphrases. We argue that since round-trip MT executes the translation two times for both directions, it is more prone to a translation error.

Our proposed scenario achieved lexically diverse paraphrases while maintaining a better semantic similarity than the round-trip translation. Alternatively, filtering the BLEU in our synthetic dataset yields to more semantically similar paraphrase but also sacrifice lexical diversity. 






 

\section{Related Work}
Prior work has shown that paraphrase can be used to provide additional data~\cite{ma2019nlpaug}, which proves to increase model performance, for example in machine translation \cite{Seraj2015,Marton2013}, question answering \cite{Dong2017}, relation extraction \cite{Zhang2015}, or text generation~\cite{gao2020paraphrase}. 
Additionally, paraphrase has been used to aid NLP evaluation~\cite{thompson2020automatic}.

There are several well-known English paraphrase corpus, such as ParaBank2 \cite{Hu2019}, PPDB \cite{Pavlick2015}, Microsoft Research Paraphrase Corpus (MSRP) \cite{Dolan2005}, Microsoft Research Video Description Corpus \cite{chen2011}, Paralex \cite{fader2013}, and Paraphrase and Semantic Similarity in Twitter (PIT) \cite{xu2015}. 
This paper compares our proposed approach with ParaBank2, which is the synthetically generated and the larger-scale corpus.

\section{Conclusion}

We proposed a way to generate a synthetic paraphrase corpus by utilizing a monolingual corpus and a translation system. The paraphrase pair is obtained by generating multiple translation samples from an English text and then pick the most diverse pair, denoted with the smallest BLEU score. With this approach, we produce a paraphrase corpus for 17 languages which we release publicly.\footnote{ https://github.com/afaji/paracotta-paraphrase} Our paraphrase is semantically similar, according to human evaluation and sBERT cosine distance evaluation. Nevertheless, our corpus is lexically diverse according to BLEU and Jaccard index.

As future work, it would be interesting to explore a different way to produce translation samples besides the beam search. Adjusting the sample size is another direction to explore, as a higher sample means that we have much more choices, therefore potentially more lexically diverse paraphrases. However, semantic similarity, as well as the computational cost required for higher sample size, must be considered. We also plan to test our approach using different NMT systems and investigate the usefulness of our dataset for downstream NLP tasks. Finally, we left for future work the details and suggestions to consider the trade-off between semantic similarity and lexical diversity.


\balance
\bibliography{main}
\bibliographystyle{acl}

\appendix

\cleardoublepage
\section{Manual Evaluation Guideline}
\label{sec:manualevalguide}
This guideline describes detailed information regarding the concept of annotation for the paraphrasing task. The paraphrase pair should be ranked by its similarity: how much the two sentences are similar semantically. The scores are defined in a 3-point-system. In this guideline, we will show some examples of each score.

\subsection{Score 3 - Completely or mostly equivalent}

\subsubsection{Equivalent meaning/Synonym}
The two sentences practically mean the same thing.
\justify
Text 1: \textit{The next morning he was found {\color{blue}unconscious}.}

\justify
Text 2: \textit{The next morning he was found {\color{blue}passed out}.}

\justify
Text 1: \textit{I {\color{blue}eat} rice.}
\justify
Text 2: \textit{Rice {\color{blue}is eaten by} me.}

\justify
Text 1: \textit{The head of the local disaster unit, Gyorgy Heizler, {\color{blue}said} the bus driver failed to notice the red light.}

\justify
Text 2: \textit{The bus driver failed to notice the red light, {\color{blue}said} Gyorgy Heizler, a head of the local disaster unit.}

\subsubsection{Identical}
Note that the exactly same sentence should be scored 3. We only care about semantic similarity. Creativity will be measured with a different scoring system.

\justify
Text 1: \textit{I eat rice}

\justify
Text 2: \textit{I eat rice}

\subsubsection{Equivalent meaning but using informal form}

Different language/style, but equivalent meaning
It is acceptable even if the text is not in formal form.

\justify
Text 1: \textit{I am {\color{blue}delighted}.}

\justify
Text 2: \textit{I am {\color{blue}chuffed}.}

\subsubsection{Generalization}
Subjects and predicates in the main clause are still equivalent or related, the case of pronoun output without additional context is considered generalization.

\justify
Text 1: \textit{Uncle has bought a {\color{blue}car}.}

\justify
Text 2: \textit{Uncle has bought a {\color{blue}vehicle}.}

\justify
Text 1: \textit{{\color{blue}Jokowi} is making a speech.}

\justify
Text 2: \textit{{\color{blue}He} is making a speech.}

\justify
Text 1: \textit{Gave a speech nine days ago in St. Petersburg.}

\justify
Text 2: \textit{{\color{blue}He} gave a speech nine days ago in St Petersburg.}

\subsubsection{Mostly Similar meaning, but there is additional/missing minor details}
\justify
Text 1: \textit{The {\color{blue}US} market is expected to fall 2.1 percent {\color{red}this year}.}

\justify
Text 2: \textit{The {\color{blue}American} market is expected to fall 2.1 percent.}

\subsubsection{Mostly Similar meaning, but differs in minor details}
\justify
Text 1: \textit{The US market is {\color{blue}expected} to fall 2.1 percent this year} 

\justify
Text 2: \textit{The {\color{blue}American} market is {\color{blue}set} to fall 2.1 percent this year.}

\subsection{Score 2 - Roughly equivalent}
An annotation score of 2 is associated with the medium similarity paraphrases: not identical but similar.

\subsubsection{Identical/mostly similar but repeated}
The two sentences were almost identical or very similar, but the output model is repeated.
\justify
Text 1: \textit{This time it was different, this time it was better.}

\justify
Text 2: \textit{This time it was different, this time it was better. {\color{red}This time it was different, this time it was better.}.}

\subsubsection{Roughly similar meaning, but there is additional/missing important information}
\justify
Text 1: \textit{Richman was irritated by Burne's tone.}

\justify
Text 2: \textit{Richman was irritated by Burne's tone. {\color{red}He felt uncomfortable with Burne's attitude}.}

\subsubsection{Roughly similar meaning, but they differ in important details}
\justify
Text 1: \textit{How long {\color{red}has it been since I last} paid you, Clifton?}

\justify
Text 2: \textit{How long {\color{red}have I} paid you, Clifton?}

\subsection{Score 1 - Inequivalent or unrelated}
\subsubsection{Not equivalent, but same topics}
\justify
Text 1: \textit{The Nasdaq composite index {\color{red}rose 10.73 or 0.7 percent to 1,514.77}.}

\justify
Text 2: \textit{The Nasdaq Composite Index, {\color{red}which is filled with technology stocks, has recently gained about 18 points}.}

\subsubsection{Very dissimilar and unrelated}
\justify
Text 1: \textit{I'm eating rice.}
\justify
Text 2: \textit{{\color{red}She fell asleep.}}

\end{document}